# Automatic Leaf Extraction from Outdoor Images

Nantheera Anantrasirichai*, Sion Hannuna and Nishan Canagarajah

*Merchant Venturers' School of Engineering, University of Bristol, UK*

*Corresponding author. Tel.: + 44 117 954 5192.

Fax:+44 117 954 5206

n.anantrasirichai@bristol.ac.uk

Automatic plant recognition and disease analysis may be streamlined by an image of a complete, isolated leaf as an initial input. Segmenting leaves from natural images is a hard problem. Cluttered and complex backgrounds: often composed of other leaves are commonplace. Furthermore, their appearance is highly dependent upon illumination and viewing perspective. In order to address these issues we propose a methodology which exploits the leaves venous systems in tandem with other low level features. Background and leaf markers are created using colour, intensity and texture. Two approaches are investigated: watershed and graph–cut and results compared. Primary-secondary vein detection and a protrusion-notch removal are applied to refine the extracted leaf. The efficacy of our approach is demonstrated against existing work.

## 1. Introduction

Plant foliage may be used to identify a particular plant's species and assess its health. Traditionally, these tasks would be carried out manually by an expert which is both time consuming and vulnerable to subjective variation. In this paper we describe and automated approach employing a variety of image processing techniques.

In our approach we isolate the leaf closest to the camera as this will not be occluded by other foliage. Once the leaf has been extracted, the margin, shape and venation can be estimated in order to identify the plant, whilst the colours of the segmented leaves can be utilised to determine the plant nutrition and health as well as the visual symptoms of plant diseases. In plant pathology diagnosis, lesion segmentation is further applied to the segmented leaf and the severity of diseases is measured as proposed by Shen et at. [1] and Kurniawati et at. [2].

Much of the previous work on leaf segmentation exploits idealised images captured in preferable scenarios: such as cuttings mounted upon homogeneous backgrounds [1]. This enables the use of traditional segmentation methods, such as thresholds, gradient operators and morphological operators. Unfortunately, these approaches are often ineffective for images captured in a natural / agricultural setting as the background typically contains partially occluded leaves, soil, branches and such like.

Fully automated leaf segmentation is an unsolved problem. Neto et al. proposed a method for automatically extracting individual leaves using the Gustafson-Kessel clustering and genetic algorithms [3]. Their approach is particularly suited to images in which the contrast between the leaf and the background is high and leaf overlap is small. Texture and colour features are used as leaves generally have regular textures whilst natural backgrounds have random textures [4]. Unfortunately the background that we want to eliminate usually includes others same-type leaves.

In this paper we exploit and compare these two well-known segmentation techniques: watershed and graph cut. Using these techniques the rough areas of objects and background are required. The higher the marker precision, the better the segmentation results. Additionally the markers have a one-to-one relationship with a specific watershed region; hence, the number of regions can be defined by the number of markers. Therefore we create two markers; one for the target leaf and the other for the background. These two markers are also used as the sink and the source seeds in the graph-cut method.

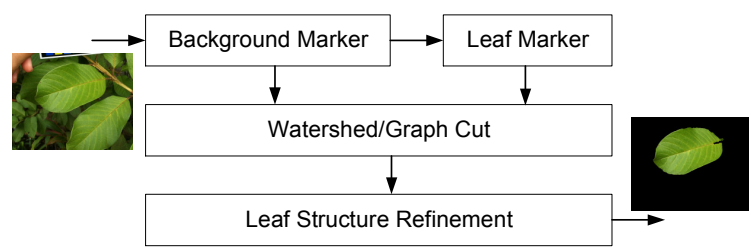

Figure 1: Leaf extraction process

The proposed framework is illustrated in Figure 1 which starts with creating the background marker and then leaf marker. The markers are then used with watershed and graph cut. Finally, the extracted leaf is refined using leaf structures. The main contributions of our approach to automated leaf extraction from natural backgrounds are as follows.

    i) ***Background marker***: We build upon existing work for colour-based background identification with a more complex algorithm which includes texture analysis and further features. This leads to better markers.

    ii) ***Leaf marker***: Not only is greyscale morphological processing used, but edge information is also exploited to ensure the marker lies in the leaf area.

    iii) ***Primary and secondary vein detection***: Veins are detected with a line detection routine, and the generally symmetric nature of leaves is exploited to identify the primary vein. This information will also be useful for plant identification and pathology [5] and, to the best knowledge of the authors, there are no methods using this information in leaf segmentation process, but it is normally detected after having an isolated leaf.

    iv) ***Protrusion and notch removal***: The contour of the extracted leaf is converted into polar coordinates. Unlikely graph shapes are detected and an estimate of the true outline is generated and used to replace the detected outline. The protrusions and notches are consequently removed.

    The rest of paper is organized as follows: Section 2 describes the related work, whilst Section 3 demonstrates the proposed automatic leaf extraction including the automatic background and foreground marker generator. The proposed leaf refinement is presented in Section 4 followed by the experimental results and discussions in Section 5. Section 6 concludes the work in this paper and the work in the future.

## 2. Related Work

Two popular segmentation methods have been used in the leaf extraction from complicated background: watershed and graph-cut. The watershed method has been used in image segmentation since the early 90's [6]. It is analogous to placing a water source in each regional minimum and then flooding the relief. Barriers are built to prevent the different sources meeting. In leaf segmentation work, a marker based watershed approach is often employed [7, 8]. Meyer and Beucher use markers in conjunction with region growing to prevent over-segmentation [9]. An Otsu threshold [10] is employed to automatically create markers [7]. This is followed by a binary erosion operation to separate other leaves from the target leaf. In order to obtain the separate markers among these leaves, either the structuring element needs to be predefined or the leaf overlap must not be too large. The watershed is applied on the gradient images of Hue, Intensity and Saturation of the HSI colour space, separately. The final result is selected from one of the three results achieving the highest solidity (integrity) measure. This method however, cannot perform well if the target leaf has strong gradients at the veins or reflection on the surface. It is not possible to achieve automatic leaf segmentation using solely colour and gradient information. In this paper, we propose a refinement process using primary and secondary vein detection, as well as a contour fitting to remove some of the protrusions and fill in the notches.

The graph-cut algorithm has been widely applied to computer vision problems that can be formulated in terms of energy minimisation. A minimal cut of the graph is achieved through the max-flow min-cut theorem [11] , where the maximum amount of flow passing from the source to the sink is computed. In image segmentation, the foreground and background seeds are used as the source and the sink as demonstrated in [12]. This can be simply applied to leaf extraction if the leaf and background are roughly known. To the best of our knowledge the graph cut has not been used to extract a leaf from a complicated background.

## 3. Leaf Extraction

The proposed background marker and leaf marker are described in this section. We assume that the processed images are 8 bits/channel/pixel, so the greyscale range is 0-255. Also, to decrease the computation time, the images having the longer length more than 600 pixels are reduced the size to 600 pixels. The positions of the extracted leaf results of the reduced-size images can straightforwardly be mapped to match those of the original images afterwards.

### 3.1 Background Marker

*Non-green background*

Non-green backgrounds, such as soil, can be simply removed using an appropriate colour histogram threshold. Following colour indices proposed in [13] an excess green index (*ExG*=2*G-R-B*) and an excess red index (*ExR*=1.4*R-G-B*), where R, G and B are red, green and blue of RGB colour space respectively, are computed.

The difference between these indices is subsequently divided into two groups according to an Otsu threshold. However if the histogram contains more than one peak, the Otsu threshold is not suitable. We therefore set the threshold at one standard deviation below the mean value if many peaks are found. Using the indices as proposed by Woebbeck et *at.* (1995) does not consider the blue component; hence we add a condition that the areas with the blue value higher than the green value ($R>G$) are also classified as background. The white ($R>200$, $G>220$, $B>200$, assuming 8 bits/channel) and black ($R, G, B <30$) areas are also defined as the background.

*Texture background*

Some areas (even those that are green) are not of interest if their textures are too highly detailed, such as grass. We utilise local entropy of greyscale values to measure statistical randomness. If the local entropy of each window (3x3) exceeds a given threshold (>220), the area is defined as the background.

*Marker cleanup*

Very small areas are possibly not background but holes in the target leaf. Areas smaller than 100 pixels and located further than 50 pixels from the nearest background area, are therefore discounted as background markers. The results of this check are eroded so as to make a small gap to the actual edge of the leaf. Finally as we are looking for the whole leaf, we mark all pixels at the edge of the image as the background.

**3.2 Leaf Marker**

The purpose of this process is to find where the target leaf is.

*Greyscale Morphology*

The inverse areas of the background markers are the inputs of this stage. The marker-based watershed segmentation is applied here. The foreground markers are found using "opening-by-reconstruction" and "closing-by-reconstruction" to clean up the image and also connect blobs of pixels inside each of the foreground objects respectively (Matworks, 2011). The local maxima of the greyscale image are calculated. The results are then forced to be the local minima in the gradient magnitude map used for watershed transform.

*Solidity measurement*

The segmented image resulting from transformation might contain many regions. The largest one that is green (i.e. the most likely to be a leaf) is initially selected. Other regions that connect to the largest region will be included in the foreground marker if they have similar intensity and hue, and if the resultant combined region has a proportion of pixels in the convex hull larger than that of the largest region.

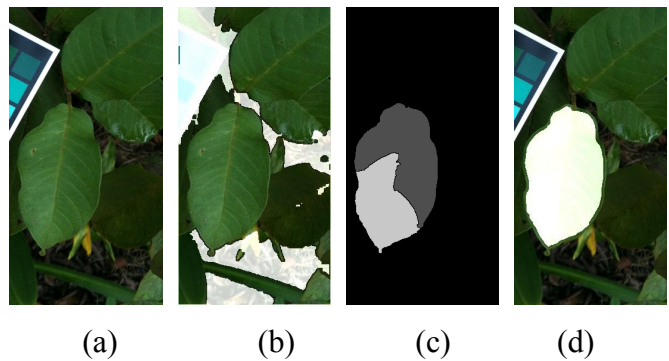

(a)      (b)      (c)      (d)

Figure 2: Marker example: (a) original Image; (b) background marker (c) two selected regions; (d) leaf marker

*Marking with Edge*

If this proportion is still smaller than a given threshold (solidity<0.95) after region-combining process, the edge information is used to cut off parts that are not in the target leaf. The detected edges connecting to the boundary of the result of the solidity measurement are used to separate the small regions that might be located outside the actual leaf. The largest area is selected and is subsequently eroded to get the front-most leaf marker. The example of the generated markers is shown in Figure 2. The more the precision of marker position, the better the segmentation result.

### 3.3 Initial Leaf

Two techniques are used to cut out the leaf region and the results of these two techniques are compared and discussed in the Results section.

**Graph cut:** The background markers and the leaf marker are used as the source seeds and the sink seeds of the process.

**Marker-controlled watershed segmentation:** The background markers and the leaf marker are used as local minima in the gradient magnitude map.

## 4. Leaf Structure Refinement

The initial leaf extraction process gives good results for most images where the boundary of the leaf shows distinct gradients; however, in some cases the target leaves exhibit an unclear boundary resulting in protrusions, and in some cases the prominent veins inside the leaf cause partially extracted leaves. Therefore we introduce using characteristics of leaves to refine the results.

### 4.1 Primary and secondary vein detection

A Hough transform [14] is employed to detect strong lines inside the boundary of the initial leaf. This facilitates checking for partially segmented leaves as a result of particularly large gradients across vein lines. The process is demonstrated in Figure 3. First, the strongest line is detected inside the initial leaf and assumed to be a primary vein. Then the secondary veins can be detected with the limited angles which are here defined to locate at the angles of 30-150 degree of the primary vein. These detected lines are subsequently used for updating the front leaf marker with the conditions as follows.

- If the strongest line detected inside the initial leaf represents the centre vein of a (mostly) symmetrical leaf and also shows some secondary veins both sides of the line and inside the initial leaf, it is assumed that the target leaf has been found thereby completing the process.

- If the strongest line detected inside the initial leaf represents the centre vein of a (mostly) symmetrical leaf but the secondary veins are not found, it implies that the initial leaf is too small. The lines outside the initial leaf are checked if they could be parted of the strongest line. Consequently the extended strongest line is included in the front leaf marker.

- If the strongest line detected inside the initial leaf doesn't represent a (mostly) symmetrical leaf, it implies that such line is not a primary vein but could be a secondary vein. The strongest lines detected at the boundary of the initial leaf are checked to see whether they are the primary vein. First the initial leaf is flipped to check if there are any secondary veins on the other side of the strongest line. If so, the lines outside the initial leaf are checked if they could be parted of the strongest line. Consequently the new detected secondary veins and the extended strongest line is included in the front leaf marker.

Finally the watershed transform is applied again. The example result of the proposed primary and secondary vein detection is illustrated in Figure 4.

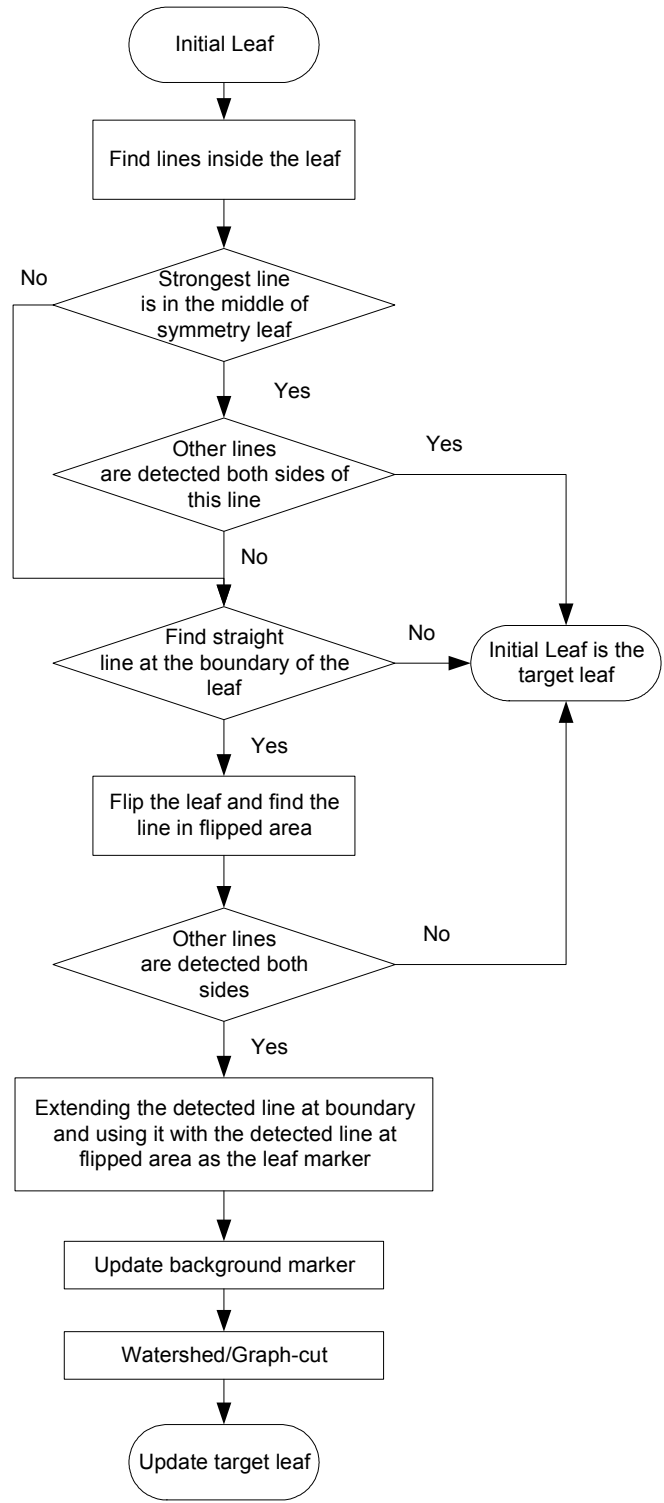

Figure 3: Process diagram of leaf refinement using primary and secondary vein detection.

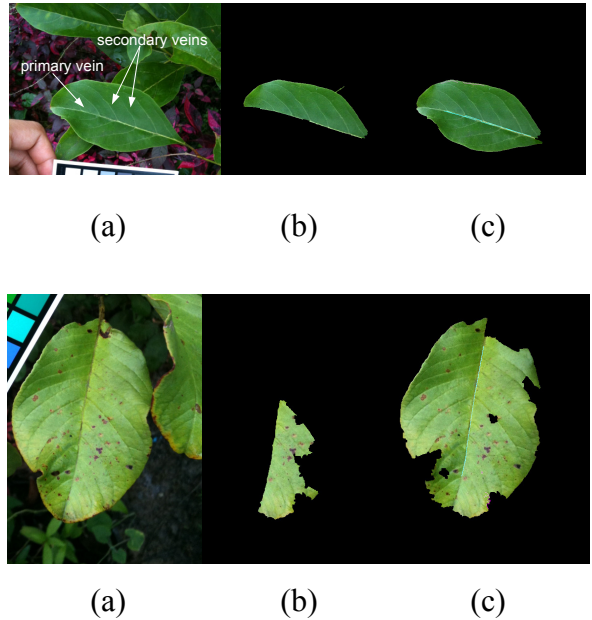

(a)      (b)      (c)

Figure 4: An example result of the proposed primary and secondary vein detection: (a) original Image; (b) result of watershed; (c) result of leaf structure refinement.

## 4.2 Leaf shape refinement

The contour of the extracted leaf is converted into polar coordinates using the centroid of the primary vein as the origin. The plot is divided into two halves (each half leaf) and operated separately. The unexpected graph shape is detected as such period of the graph show high varying values. Subsequently the coefficients of a polynomial of degree of 3 ($y = ax^3+bx^2+cx+d$) are estimated to fit to the smooth data in a least squares sense. The unexpected plot is then replaced with the estimated fit curve. Consequently the protrusions are removed and the notches are filled. The example of the refinement results is demonstrated in Figure 5.

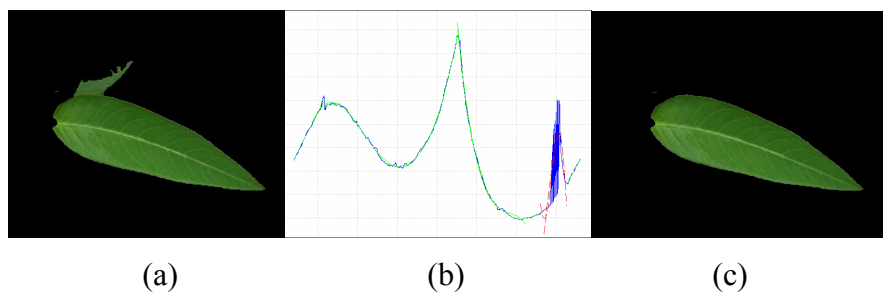

(a)      (b)      (c)

Figure 5: Protrusion and notch removal using curve fitting: (a) results of initial extracted leaf; (b) contour curve with protrusion mark; (c) refined result.

## 5. Results and Discusstion

The proposed method is tested with 100 random leaf images and also compared to those in [8]. Note that the images having the longer length more than 600 pixels are reduced the size to 600 pixels so as to decrease the computation time.

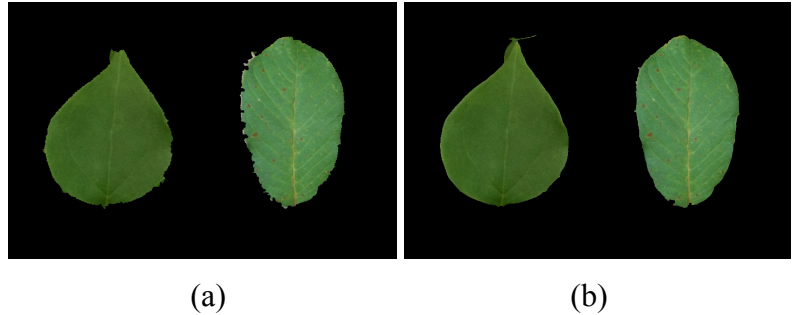

(a)           (b)

Figure 6: shows the cut-out leaves: (a) graph cut; (b) watershed

Two popular methods of image cut out, watershed and graph-cut, were exploited and compared. The results for watershed and graph-cut are reasonably similar. Graph-cut achieves better results when the leaf exhibits high gradients in the veins, whilst the watershed gives a smoother edge. The examples are illustrated in Figure 6.

We evaluate the final results with the ground truth obtained from manual segmentation. The precision, $P$, and recall, $R$, measurement is used as shown in equations below.

$$P = |L \cap G| / |L| \qquad (1)$$

$$R = |L \cap G| / |G| \qquad (2)$$

where $L$ and $G$ are the sets of pixels representing the leaf for the segmentation algorithm and ground truth respectively. Precision represents the proportion of the pixels labelled as leaf that are correct, whilst recall represents the proportion of the leaf that has been labelled as such. Both precision and recall close to 1 represents good segmentation.

The precision-recall plot is illustrated in Figure 7 and the examples of subjective results are shown in Figure 8. Note that the results shown in those figures were generated using watershed transform. The average precision and recall of our scheme are 0.92 and 0.90 respectively, whilst those of Tang et at (2009) are 0.78 and 0.75, respectively. Tang's scheme has cases with precision and recall both zero, because some extracted results are entirely incorrect (Figure 8 (c)) as their method is heavily relied on image intensity and Otsu threshold. Our approach doesn't have similar problems because we have an improved threshold method and their method makes assumptions about the input image that ours doesn't require. If the images with the precision and recall of zeros are not included, the average precision and recall of our scheme are 0.92 and 0.91 respectively, whilst those of Tang' scheme are 0.83 and 0.80, respectively. We also evaluate the performance of the proposed structure leaf refinement by comparing the results of the initial extracted leaves to the final extracted leaves. The average precision and recall of the initial leaf process are 0.86 and 0.84 respectively which are 0.06 less than the final results. It proves that the refinement process can improve the extraction performance and it also shows that our proposed leaf and background markers outperform those of Tang' scheme.

Figure 8 reveals that our scheme still misses some areas (low recall value). This occurs because of reflection, shadow and disease on the leaves. The results with various types of leaves and different backgrounds are shown in Figure 9.

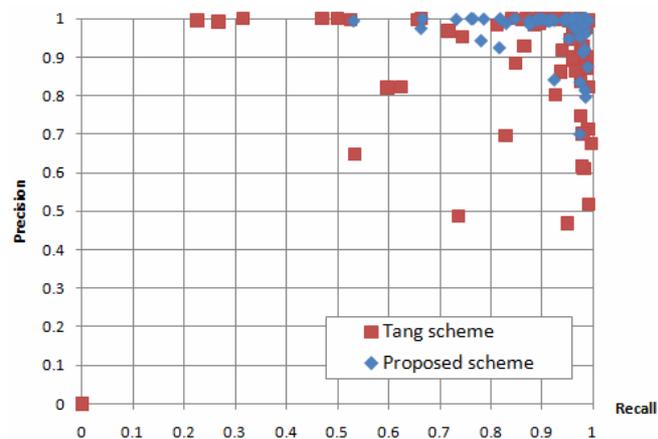

Figure 7: Precision-recall shows region extraction performance and comparison between our proposed scheme and the existing scheme proposed by Tang et at. [8]

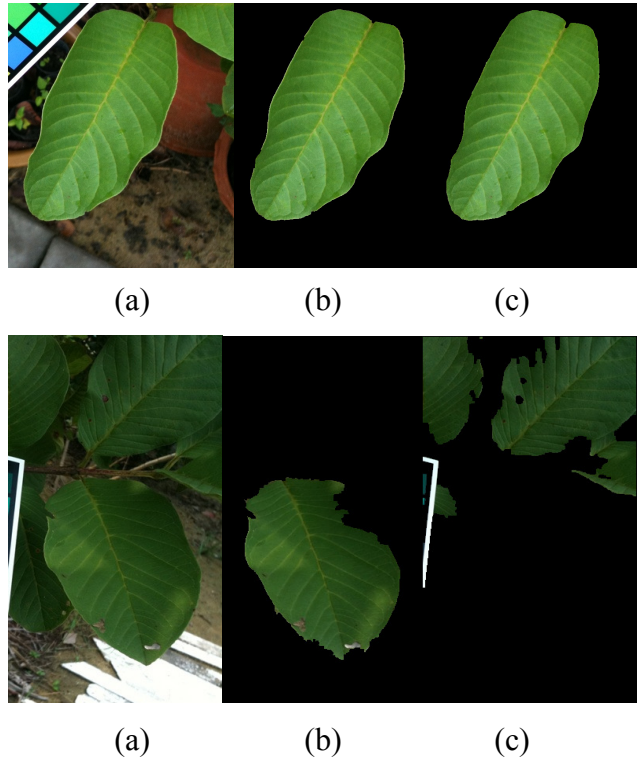

(a) (b) (c)

(a) (b) (c)

Figure 8: Example leaves: (a) original image; (b) proposed scheme and (c) Tang's scheme

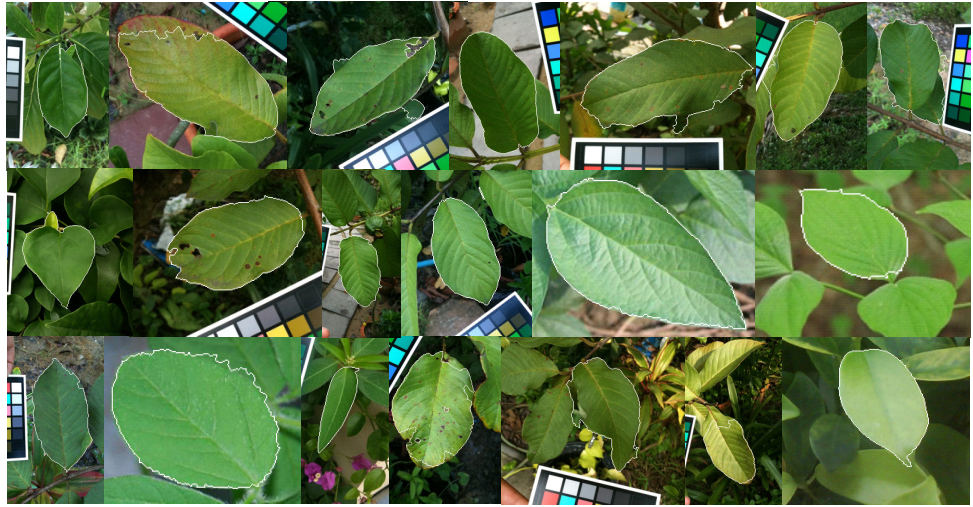

Figure 9: Examples of the results of the proposed scheme

## 6. Conclusions

In this paper, we propose an automatic process for extracting a leaf from a natural background. The process includes an automatic background marker and an automatic leaf marker which will be used in the image cut-out technique. This paper explores two techniques: watershed and graph-cut. To obtain the final result we propose a leaf refinement using leaf characteristics – colours, shape and venation. The experiment shows better results than the existing technique. In the future we will develop a lesion segmentation and an automatic leaf-disease classification which is one of the aims of the project.

# References

[1] Shen W., Wu Y., Chen Z and Wei H., Grading Mothod of Leaf Spot Disease based on Image Processing, in Proc. Conf. Computer Science and Software Engineer 6, 2008, pp.491-494.

[2] Kurniawati N.N., Abdullah S.N.H.S. and Abdullah S., Investigation on Image Processing Technique for Diagnosing Paddy Diseases. In Proc. Conf. Soft Computing and Pattern Recognition, 2009, pp. 272-277.

[3] Neto J., Meyer G., Jones D., and Samal A., Plant species identification using Elliptic Fourier leaf shape analysis. Computers and Electronics in Agriculture 50, 2006, pp. 121–134.

[4] Liu W. and He Y., Fractal features and fuzzy clustering based tree leaf segmentation from images with the complex background, In Proc. International Conference on Computer, Mechatronics, Control and Electronic Engineering, 2010, pp. 317-321.

[5] Fu H. and Chi Z., Combined thresholding and neural network approach for vein pattern extraction from leaf images. In Proc. Vision Image and Signal Processing 153(6), 2006, pp. 881-892.

[6] Meyer F., Topographic distance and watershed lines. Signal Processing 38, 1994, pp. 113-125.

[7] Wang X.F., Huang D.S., Du J.X., Xu H. and Heutte L., Classification of Plant Leaf Images with Complicated Background. Journal on Applied Mathematics and Computation 205, 2008, pp. 916-926.

[8] Tang X., Liu M., Zhao H. and Tao W., Leaf Extraction from Complicated Background. Proceedings of IEEE International Congress on Image and Signal Processing, 2009, pp. 1-5.

[9] Meyer F. and Beucher S., Morphological segmentation. Journal on Visual Commun. Image Represent 1(1), 1990, pp. 21-46.

[10] Otsu N., A Threshold Selection Method from Gray-Level Histograms. IEEE Transactions on Systems, Man, and Cybernetics. 9(1), 1979, pp. 62-66.

[11] Boykov Y., Veksler O. and Zabih R., Fast approximate energy minimisation via graph cuts. IEEE Transactions on Pattern Analysis and Machine Intelligence 29, 2001, pp. 1222–1239.

[12] Boykov Y. and Funka-Lea G., Graph cuts and Efficient N-D Image Segmentation. International Journal of Computer Vision 70(2), 2006, pp. 109-131.

[13] Woebbeck D.M., Meyer G.E., Von Bargen K. and Mortensen D.A., 1995. Colour Indices for Weed Identification under Soil, Residual and Lighting Conditions. Trans. ASAE 38, 1995, pp. 259-269.

[14] Gonzalez R.C., Woods R. E. and Eddins S. L., Digital Image Processing Using MATLAB. Prentice-Hall, 2003.